# Randomized fast no-loss expert system to play tic tac toe like a human

*Aditya Jyoti Paul*[1,2]
1. Cognitive Applications Research Lab, India.
2. Department of Computer Science and Engineering, SRM Institute of Science and Technology, Kattankulathur, TN – 603203, India.
*Email:* aditya_jyoti@srmuniv.edu.in

**Abstract:** This paper introduces a blazingly fast, no-loss expert system for Tic Tac Toe using Decision Trees called T3DT, that tries to emulate human gameplay as closely as possible. It does not make use of any brute force, minimax or evolutionary techniques, but is still always unbeatable. In order to make the gameplay more human-like, randomization is prioritized and T3DT randomly chooses one of the multiple optimal moves at each step. Since it does not need to analyse the complete game tree at any point, T3DT is exceptionally faster than any brute force or minimax algorithm, this has been shown theoretically as well as empirically from clock-time analyses in this paper. T3DT also doesn't need the data sets or the time to train an evolutionary model, making it a practical no-loss approach to play Tic Tac Toe.

## 1. Introduction

Tic Tac Toe is one of the oldest and most played childhood games, dating back as early as 1300BC in Egypt. The board comprises of a 3x3 matrix; two players alternatively place an X or an O in an empty cell and the first player to get three X's or O's in a row, column or diagonal wins. It is a zero-sum game in which one player's win is the other's loss. It also is a full information game, that is both the players have complete information about all the moves that are possible arising from a certain board state. Over the ages, people have come up with tricks or techniques to play it without being beaten. The operations in decreasing priority order are:
    *a.* making a winning move, else
    *b.* blocking the opponent's winning move, else
    *c.* trying to make a fork, and lastly
    *d.* blocking an opponent's fork.

The main objective of this research work is the creation of a no-loss expert system that can play Tic Tac Toe by understanding the rules of the game and maintain a high win-to-draw ratio. Here no-loss refers to the game never losing against an opponent, it would draw or preferably win the game, irrespective of whether it starts first or second. It should be able to execute of the operations listed above just like a human would, so a decision-tree based algorithm (T3DT) was chosen. Humans' perception of games is fuzzy and chaotic in nature and the randomness hence produced makes them fun to be played with. T3DT tries to recreate the randomness to the maximum extent possible, thus making the algorithm a fun opponent to play with.

## 2. Literature review

Many heuristics and mathematical algorithms have been designed to win the game. In the field of two-player zero-sum games, pioneering work was done by John von Neumann, his first proof [1] of the minimax algorithm published in 1928 created ripples through the scientific and mathematical communities and is still considered one of the founding blocks of game theory. Kjeldsen [2] iterated the history of the development of Neumann's minimax from his paper [1] in 1928 to his completely different proof in his book [3] with Morgenstern in 1944. In [2], Kjeldsen wrote that by considering mixed strategies, and expressing player values in the bilinear form h, Neumann [1] had shown that for two-player zero-sum games, there always exists optimal mixed strategies $\xi_0$, $\eta_0$, such that

$$max_\xi \, min_\eta \, h(\xi, \eta) = min_\eta \, max_\xi \, h(\xi, \eta) = h(\xi, \eta) \qquad (1)$$

Kjeldsen [2] wrote that Neuman had actually proven a generalized version of minimax considering a class of functions broader than the bilinear form h. However, for simpler two-player zero-sum games like Tic Tac Toe, the essence of this equation lies in the observation that the optimal strategy for both players involves pessimistically minimizing the maximum damage that can be inflicted by the opponent.

Minimax surprised people by providing an algorithm which can never be beaten, and most of the earlier approaches for Tic Tac Toe relied on some form of game tree search. Claude Shannon realized the importance of variable depth search in 1949 [4], and best-first approaches with both fixed [5] and variable depth [6] have been implemented since. In general, much of the early research on game-tree search algorithms like alpha-beta pruning [7-9], Scout [10], NegaScout [11], SSS* [12-15], fixed and dynamic node ordering [16] and aspiration windows [17-18] make the same choice as full-width fixed depth minimax. On the other hand, algorithms like B* [19-20], min-max approximation [21], conspiracy search [22-23], meta-greedy search [24], singular extensions [25], and risk assessment [26] might not always do so, searching some strategies more deeply than others. Improved hardware and algorithms supporting parallelism [27-30] have made tree generation and search even faster. Here, just a cursory description of game tree search methods has been given as a detailed account is outside the scope of this paper, but [31-33] compare and contrast these methods in much greater detail for a clearer understanding.

Ignoring symmetry of the board, there are 255,168 unique games of Tic Tac Toe [34], and move-generation and analysis of all these moves at runtime is tedious and time-consuming. Humans are skilled at simplifying the search process by pruning the options at hand by having an intrinsic understanding of the game and selecting a few main candidates [35]. This task is challenging for computers and thus other approaches which do not rely on game tree search were developed and have been described below.

Fogel [36] laid the groundwork for using evolution in developing neural agents to play Tic Tac Toe. Chellapilla and Fogel [37] demonstrated how intelligence is developed in evolutionary neural networks, and how they can learn to play any arbitrary game without depending upon human expertise or being explicitly programmed to do so. Hochmuth [38] described how this game could be played by a genetic algorithm. Soedarmadji [39] suggested a decentralized approach comprising of 9 agents, one for each cell in the board and one manager to choose which move to do make based on the priority number generated by each independent agent. Y. Yau et al. [40] compared the implementations of various adaptations and related parameters in evolving agents to play games like Tic Tac Toe. This paper [41] comprehensively discussed the use of co-evolution and pareto evolution in development of neural agents and how success is strongly dependent on the initialization of– the co-evolution process, and related empirical data is discussed.

Bhatt et al. [42] did some commendable work, trying to find out no-loss strategies for the game using a customized genetic algorithm, focusing on a high win-to-draw ratio. They succeeded in finding 72,657 such strategies, and also came to some other interesting conclusions, which verify our existing notions about the game. Mohammadi et al. [43] presented a co-evolution and interactive fitness based genetic algorithm to build a human-competitive player. Ling et al. [44] used double transfer function and Rajani et al. [45] applied the Hamming distance classifier, on neural networks to demonstrate the advantages of each method. A. Singh et al. [46] devised a deterministic mathematical model to solve Tic Tac Toe as a nine-dimensional problem. Karamchandani et al. [47] discussed the techniques used in [44-45] and presented a very basic rule-based algorithm, which always targets the centre at the start and has a set of special moves, but it can still be beaten by an opponent in some cases. Sneha Garg et al. [48-49] formally defined the Tic Tac Toe problem and formulated a winning strategy for the same using a multi-tape Turing machine and automata theory.

Initial approaches by Hochmuth [38] and Soedarmadji [39] were innovative but not very effective. Hochmuth's method does not guarantee no loss for all possible tic-tac toe games, and was an exercise demonstrating the effectiveness of genetic algorithms, rather than the development of a perfect no-loss algorithm. Soedarmadji's heuristic approach loses in at least three board states as described in this paper [42]. The three examples are explained through Fig. 2.1 below, in which, optimal positions for X are marked with '+'. In the first example, X should have marked any of the edge-centres, not in the top-right corner which opens a fork for the opponent, the second example is more evident as X fails to block a two

O's in a row and similarly in the third example X again fails to block O.

It is discussed in [40] and [42] and in Sec. 4 of this paper itself that the optimal strategy for the first and second player are different from each other. The first player has a greater chance of victory (around 1.68:1 [42]). In [50], a non-randomized decision tree approach is discussed by Sriram et al. They demonstrated some interesting observations, like how minimax fails to move optimally in as many as 25-40 cases. However, the algorithm was unclear in some aspects, like it was not discussed how the algorithm would make a move if it had to start the game itself, even though the strategies are different for the two players as discussed above. And when the opponent makes the first move, it was suggested to always mark the centre if possible else mark the top-left corner, according to the tree in Fig. 2.2.a, ignoring the edges completely. The remaining moves were to follow the tree in Fig. 2.2.b. Further it was unclear exactly how each of the steps in the decision tree were carried out as that would have a significant effect on the actual runtime of the algorithm. Runtime comparison against vanilla minimax was done solely on the basis of asymptotic complexity and not clock time complexity. Asymptotic complexity does not give a true representation

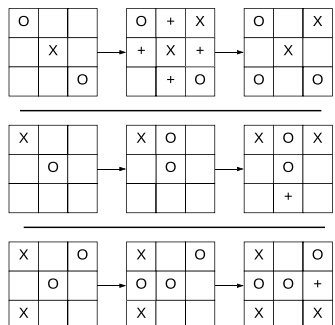

***Fig. 2.1***: *Three examples in which Soedarmadji's solution [39], playing with X, loses the game [42].*

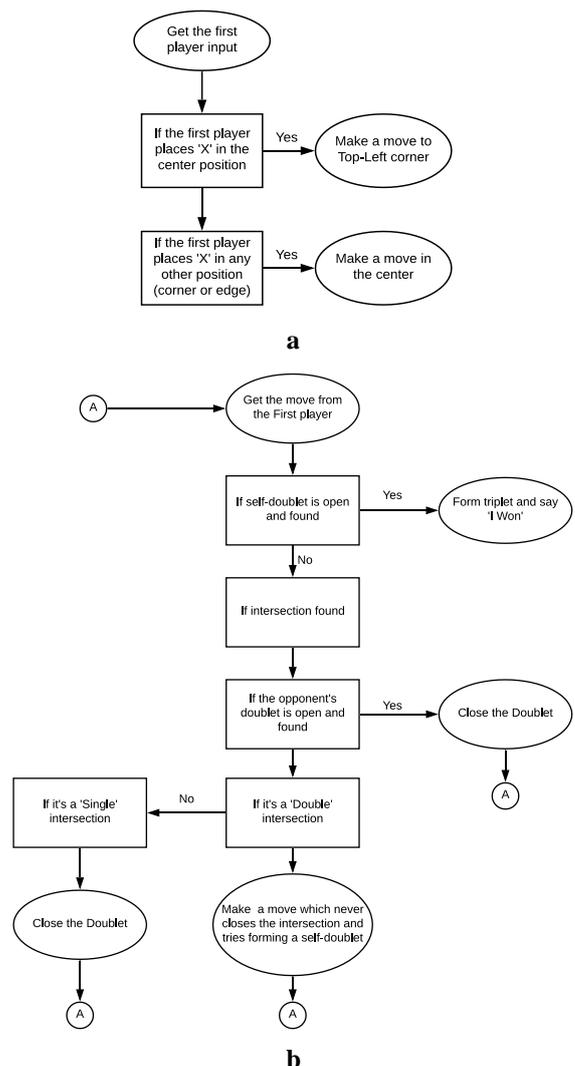

***Fig. 2.2***: *Decision trees for the algorithm suggested in the paper [50]. **(a)** Decision tree for the first move, **(b)** Decision tree for the remaining moves*

of the time taken by an algorithm to run, which is needed for benchmarking. All these shortcomings are addressed in this research work.

Thus, it's observed that initially most of the game-playing algorithms were inspired by minimax and speed-ups involving pruning the game tree by various methods. Gradually approaches based on genetic and evolutionary algorithms were proposed, and these can generalize well to larger boards, but they take time to train, and needs a careful tweaking of the hyperparameters. It is also difficult to estimate how many of these no-loss strategies can actually win against a sub-optimal player, which is a necessary trait to make the algorithm more human-like, for example, the 'forgetting problem' described in [41] prevents 'expert level' players from winning or even drawing against medium level or random players.

People have through the ages, tried to intuitively come up with rules to win the game. However, surprisingly not much research has been done to either try to design a randomized rule-based solution until now, that tries to maximize win-to draw ratio for Tic Tac Toe and compare such an algorithm's clock time complexity to minimax and similar tree search approaches. Although most of the techniques to play Tic Tac Toe are centuries old, [51-52] reported some rules which could be incorporated into an expert system like T3DT. However, this book [51] steers clear from starting in an edge, which was necessary for the randomization and hence incorporated into T3DT. In [52], Ryan Aycock went to great lengths to explain the motivation for some of the rules for effective gameplay.

Hence, this research work proposes a novel no-loss expert system to play tic tac toe, called T3DT and compares the performance both theoretically as well as from actual runtime to existing minimax-based procedures, proving that it is faster and a practical approach to play tic tac toe. The rest of the paper is organized as follows. In Sec. 3, the optimality of some existing minimax-based tree-search algorithms, which have been used for benchmarking, are briefly discussed. In Sec. 4, the proposed algorithm is explained in detail. In Sec. 5, the experimental methodology is described and some newly devised metrics for the benchmarking process are elucidated. In Sec. 6, the algorithm is compared to the existing algorithms and the results are compared and analysed. Finally, Sec. 7 gives the conclusion and discusses some avenues for further work.

## 3. Optimality of Some Existing Tree-search Algorithms

This section gives a brief overview and discusses the optimality of three commonly used minimax-based no-loss strategies, namely:
    **a.** Minimax (MM)
    **b.** Minimax with alpha beta pruning (ABP)
    **c.** Minimax with advanced alpha beta (ABA)

Optimality refers to the algorithm arriving at the goal in the least number of moves. Here the vanilla minimax (MM) and minimax with alpha beta pruning (ABP) have their usual implementations. Minimax with advanced alpha beta (ABA) has a modified score function which subtracts the number of moves left till victory from the returned score. The vanilla minimax algorithm and ABP never lose but they may occasionally make a move that results in a slower victory. For example, the opponent starts the game and after both players alternatingly making moves $O_1$, $X_1$, $O_2$, $X_2$ and $O_3$, the bot's $X_3$ moves chosen by each of the algorithms are shown in Fig. 3.1. Here, marking $X_3$ in cell (3,1) would result in a victory on the diagonal instantly. MM and ABP do not choose this move, they still win eventually, but take a longer path. Including the depth into the evaluation function allows ABA to pick the optimal winning move (3,1) like T3DT leading to the fastest victory.

| $O_1$ | $O_2$ | $X_2$ | $O_1$ | $O_2$ | $X_2$ | $O_1$ | $O_2$ | $X_2$ |
|---|---|---|---|---|---|---|---|---|
|  | $X_1$ |  | $X_3$ | $X_1$ |  |  | $X_1$ |  |
| $O_3$ |  | $X_3$ |  | $O_3$ |  |  | $X_3$ | $O_3$ |

        *a*                *b*                *c*

***Fig. 3.1***: *Different moves for $X_3$ made by the algorithms MM (**a**), ABP (**b**) and ABA (**c**) on the same board state. T3DT makes the same winning move as ABA in this state.*

## 4. Proposed Method

In this section, the no-loss system called T3DT is detailed, which is much faster than minimax-based procedures, because it logically partitions the entire game into small subparts that are easier to analyse. Henceforth the computer or AI player will be referred to as the 'Bot', and it is assumed for simplicity that the bot always plays with X and the opponent plays with O. The algorithm can be divided into 2 strategies:
   a) When the bot starts the game (Sec. 4.1 - 4.2) and
   b) when the opponent starts the game (Sec. 4.3 - 4.4)

### 4.1. Algorithm when the Bot starts the game

**Start**
**Step 1**: *The Bot makes the first move, with absolute randomness in a corner edge or centre.*
**Step 2**: *The opponent marks any of the remaining cells.*
**Step 3**: *Depending on the opponent's first move in Step 2, the Bot makes its second move as shown in Fig. 4.1, and explained in Sec. 4.2,*
**Step 4**: *Then the Opponent makes its second move.*
**Step 5**: *Disregarding the Opponent's second move, the Bot always tries to first make a winning triplet or block the Opponent, if no such possibilities exist, it makes use of the strategies shown in Fig. 4.1, and explained in Sec. 4.2.*
**Step 6**: *The Opponent makes its move.*
**Step 7**: *The Bot makes its consequent moves, based on the strategies mentioned in section 4.5, which comprise of winning or blocking or making a random move.*
**Step 8**: *Repeat steps 6 and 7 till the game ends.*
**Step 9**: *Declare the result.*
**End**

### 4.2. The First Few Moves when the Bot starts the Game

When the bot starts the game, to make the game truly randomized, it chooses corner, edge or centre with equal probability. These three cases again have their own sub-parts, which are all described in Fig. 4.1.

In the following subsections 4.2.1 – 4.2.3, the motivation for each of the choices made in this decision tree are elaborated.

### 4.2.1. Bot Starts in a Corner

The Opponent chooses one of the following positions, based on which the Bot makes the next few moves:

**a. Edge**: If the opponent chooses an edge, then the bot chooses the centre as its second move. Now since the bot has made 2 moves along a diagonal, the opponent is forced to mark the other corner of the diagonal to block the bot from winning. Now unless it needs to win or block the opponent, the bot marks a corner in its 3rd move, such that it creates an empty V shaped fork, this guarantees a win. In both of the games in Fig. 4.2, irrespective of position of $O_3$, the Bot always wins.

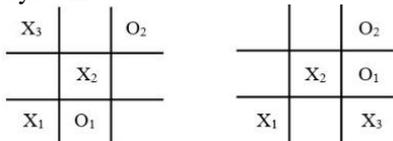

*Fig. 4.2*: *Bot starts in corner and opponent's first move is in an edge*

**b. Corner**: If the opponent chooses another corner, then the bot marks any of the remaining free corners. The opponent is now forced to block the bot's win and the bot marks the last free corner, thus creating a fork for itself. Once again this guarantees a win for the bot, like in the games in Fig. 4.3.

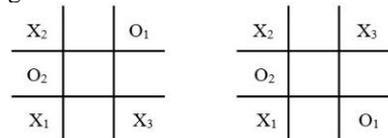

*Fig. 4.3*: *Bot starts in corner and opponent's first move is in a corner*

**c. Centre:** Suppose the opponent marks the centre, then the bot marks the diagonally opposite corner to the first X. Now if the opponent places $O_2$ another corner, the bot's win is guaranteed (see resulting board on left in Fig. 4.4); else the game will end in a draw (see resulting board on right in Fig. 4.4) for further optimal play.

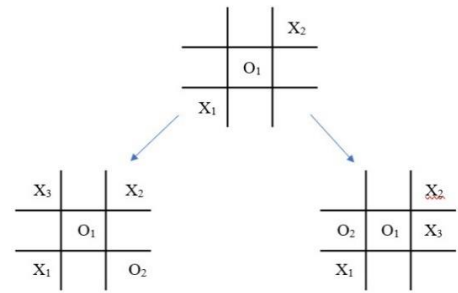

*Fig. 4.4*: *Bot starts in corner and opponent's first move is in the centre*

### 4.2.2. Bot Starts in an Edge

This is the most complicated and interesting part of the algorithm, when the bot chooses an edge. It would be much simpler to just prevent the bot from starting in the edge, as suggested by multiple sources like [51-52], but to make the game randomized, this starting move has to be included. The opponent may choose any of the following cases, based on which the Bot's subsequent plays are explored:

**a. Corner**: If the opponent chooses a corner, the bot moves in the centre and in the third move, it marks any of the remaining empty edges, unless a winning or blocking opportunity arises. This move-sequence almost always results in a draw, like the two games in Fig. 4.5.

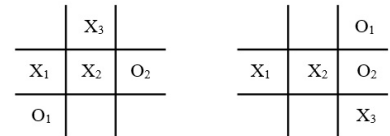

*Fig. 4.5*: *Bot starts in edge and opponent's first move is in a corner*

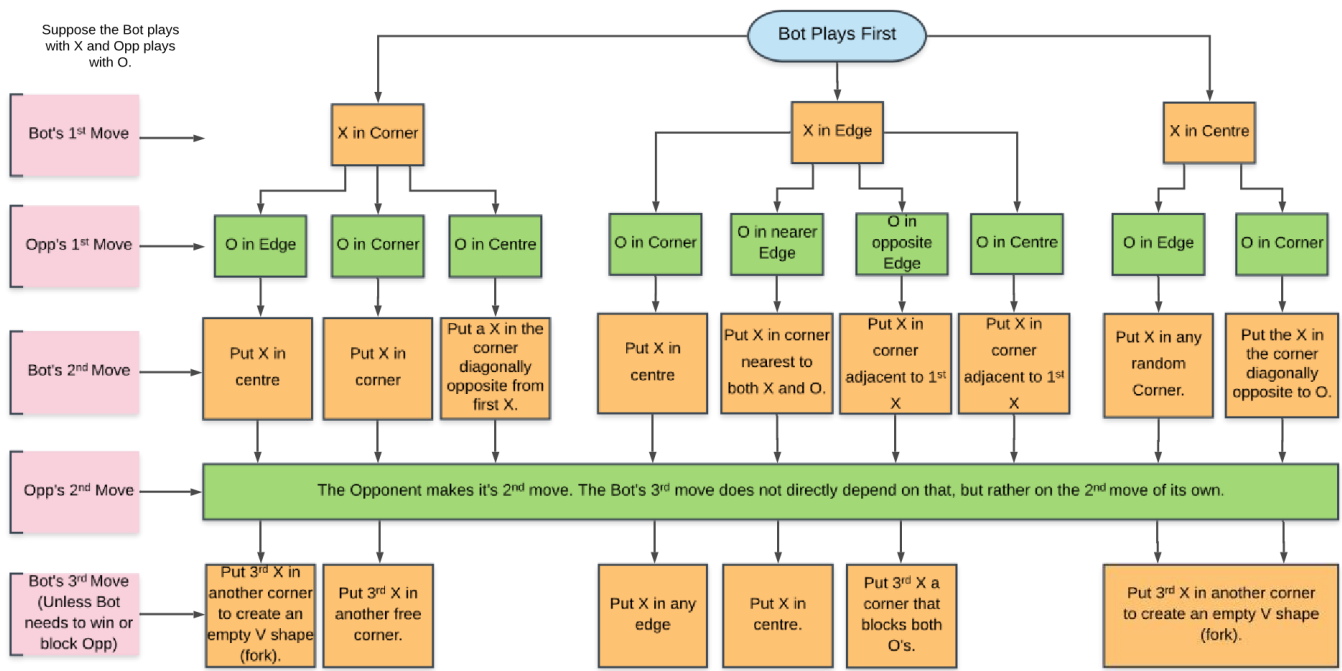

*Fig. 4.1*: *Decision tree when the Bot starts the game.*

**b. Edge:** It is again of two types:

**i. Edge Nearer to the Bot's first move:** Here we will attempt to create an L-shaped fork making sure the bot wins. If the opponent chooses any of the two edges nearest to the bot's chosen edge, the bot puts a $X_2$ in the square adjacent to both the bot's $X_1$ and opponent's $O_1$. Now the opponent will be forced to block the bot and the bot will now make its third move I the centre. This sequence guarantees a win for the bot, because of the L shaped fork that is formed, as seen in Fig. 4.6.

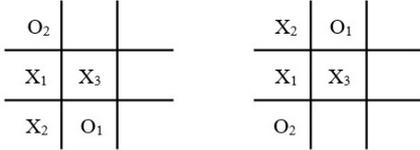

*Fig. 4.6*: *Bot starts in edge and opponent's first move is in nearer edge*

**ii. Edge opposite to the Bot's first move:** If the opponent chooses the opposite edge to the bot's first move $X_1$, the bot places a cross in any of the corners adjacent to $X_1$. The opponent will block the bot with $O_2$ in corner. Now the bot places $X_3$ in the corner that is closest to the opponent's moves $O_1$ and $O_2$. This sequence results in a draw, as in Fig. 4.7.

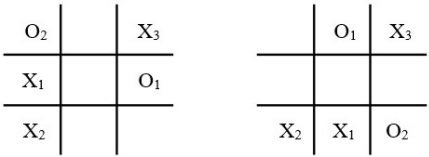

*Fig. 4.7*: *Bot starts in edge and opponent's first move is in opposite edge*

**c. Centre**: If the opponent chooses a centre, bot marks the corner adjacent to its first move. This always ends in a draw for optimal play, as in Fig. 4.8.

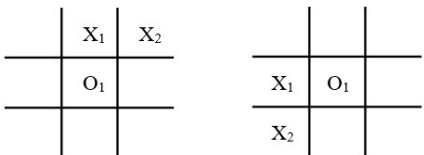

*Fig. 4.8*: *Bot starts in edge and opponent's first move is in centre*

### 4.2.3. Bot starts in the Centre

Then the opponent chooses either of the following positions, based on which the Bot makes its subsequent moves:

**a. Edge**: If the opponent chooses an edge, the bot chooses any of the 4 corners randomly. Now the opponent needs to block the bot by its move $O_2$. The bot takes advantage of this and creates a fork which is an empty V-shape, by placing $X_3$ in the required corner. This sequence guarantees victory for the bot, as in both the games in Fig. 4.9

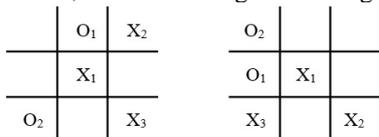

*Fig. 4.9*: *Bot starts in the centre and opponent's first move is in an edge*

**b. Corner**: If the opponent chooses a corner, the bot marks the diagonally opposite corner. Now unless the opponent moves in a corner, the bot will always win, by forming an empty V-shaped fork, as in Fig. 4.10.

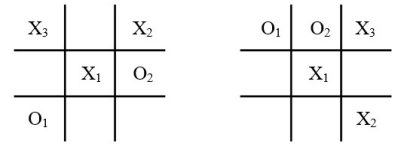

*Fig. 4.10*: *Bot starts in the centre and opponent's first move is in a corner*

### 4.3. Algorithm when the Opponent starts the Game

**Start**
**Step 1:** *Bot waits for the Opponent to make its first move.*
**Step 2:** *Based on Step 1, the Bot makes its first move as shown in Fig. 4.11, and explained in Sec. 4.4,*
**Step 3:** *Then the Opponent makes its second move.*
**Step 4:** *The Bot always tries to first make a winning triplet or block the Opponent, if no such possibilities exist, then it makes its move depending on the strategies shown in Fig. 4.11, and explained in Sec. 4.4.*
**Step 5:** *The Opponent makes its move.*
**Step 6:** *The Bot makes its consequent moves, based on the strategies mentioned in Sec. 4.5, which comprise of winning or blocking or making a random move.*
**Step 7:** *Repeat steps 6 and 7 till the game ends.*
**Step 8:** *Declare the result.*
**End**

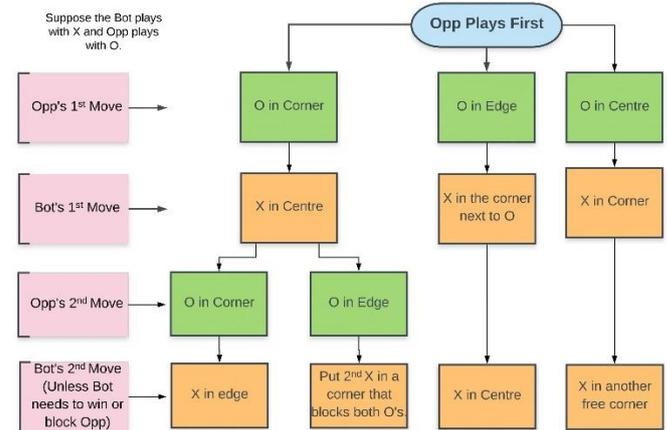

*Fig. 4.11*: *Decision tree when the Opponent starts the game.*

### 4.4. The First Few Moves when the Opponent starts the Game

The Opponent's first move may be any position on the board – a corner, an edge or the centre. This leads to three strategies for the Bot, which are all described in Fig. 4.11.

#### 4.4.1. Opponent starts in a Corner

If the opponent starts in corner, the bot moves in centre as its first move. Now the opponent's second move may be either a corner or an edge, based on which the Bot makes its subsequent moves.

**a. Corner**: If opponent's second move is a corner, bot's second X is in any of the four empty edges, unless there is a need to block the opponent. In other words, if the opponent marks the corner diagonally opposite to the first move, then the Bot marks a random edge, as seen in the left board in Fig. 4.12, else the bot will need to block the opponent, like in the right board in Fig. 4.12. These move sequences usually result in a draw.

|     |     |     |
|-----|-----|-----|
| O₁  |     |     |
|     | X₁  |     |
|     | X₂  | O₂  |
| O₁  |     |     |
|     | X₂  | X₁  |
|     |     | O₂  |

*Fig. 4.12*: *Opponent starts in the corner and opponent's second move is in another corner*

**b. Edge**: If opponent's second move is in an edge, unless the bot needs to block the opponent, its second $X_2$ is in the corner that blocks both the O's. In other words, $X_2$ is in the corner which is at a minimum distance from $O_1$ and $O_2$. These move sequences usually result in a draw, as in Fig. 4.13.

|     |     |     |
|-----|-----|-----|
| X₂  | O₂  |     |
|     | X₁  |     |
| O₁  |     |     |
|     |     | O₁  |
|     | X₁  |     |
|     | O₂  | X₂  |

*Fig. 4.13*: *Opponent starts in the corner and opponent's second move is in an edge*

### 4.4.2. Opponent Starts in an Edge

When the opponent starts in an edge, the bot marks any of the two corners next to opponent's first move $O_1$, and unless there is a need to block the opponent, which arises only if the opponent has marked $O_2$ in the centre, the bot puts second move $X_2$ in the centre. These move sequences usually result in a draw, like in Fig. 4.14.

|     |     |     |
|-----|-----|-----|
| X₁  | O₁  |     |
|     | O₂  |     |
|     | X₂  |     |
|     | O₁  | X₁  |
|     | X₂  |     |
|     | O₂  |     |

*Fig. 4.14*: *Opponent starts in an edge*

### 4.4.3. Opponent starts in the centre

If the opponent starts in the centre, the bot's first move is in a corner, and if there's no need to block (that is if the opponent marks the cell diagonally opposite to bot's first move), the second bot's move is placed randomly in any of the two remaining corners. This sequence of moves usually leads to a draw, line in the boards in Fig. 4.15.

|     |     |     |
|-----|-----|-----|
|     | O₂  |     |
|     | O₁  |     |
| X₁  | X₂  |     |
|     |     | X₂  |
|     | O₁  |     |
| X₁  |     | O₂  |

*Fig. 4.15*: *Opponent starts in the centre*

### 4.5. The Last few moves

In all of the above, irrespective of whether the bot or the opponent starts the game, after the decision tree has played its part, the rest of the bot's moves comprise of the three tasks in the following priority order:
   a. Make a triplet and win,
   b. Block the opponent's triplet,
   c. Randomly select any one of the empty squares; because now there is no longer a way to win (or lose) the game, and the moves are kept randomized in order to add some variety.

## 5. Experimental Methodology and Proposed Comparison Metrics

This section discusses the implementational details, experimental setup and the newly formulated notations for this research work. The algorithm is implemented in Java 8. All times are measured in nanoseconds, using the internal nanosecond counter called by the method nanoTime() of System class [53], which has a high degree of precision. Runtimes were calculated for the four algorithms MM, ABA, ABP and T3DT (please refer to Sec. 3 and 4 for a description of these algorithms).

For these comparisons, each algorithm was made to play against itself thousands of times, on Java HotSpot(TM) 64-Bit Server VM, with the JIT compiler both enabled and disabled, on multiple computers having dissimilar specifications to simulate all the possibilities and these statistics have been tabulated. JIT compiler was disabled to prevent the runtime optimization for this benchmarking. Always the runtimes or the first fifty games were dropped, as they might have some noise due to the cold start of the compiler. Three of these setups whose findings would be presented, have their specifications listed in Table 5.1.

*Table 5.1*: *Specifications of setups used for benchmarking*

| Specifications | Setup 1 | Setup 2 | Setup 3 |
|---|---|---|---|
| Processor | Intel Core™ i5-7200U | Intel Core™ i5-7200U | AMD A4-3330MX APU |
| Cores/ Threads | 2/4 | 2/4 | 2/2 |
| Base Frequency | 2.5Ghz | 2.5Ghz | 2.2Ghz |
| Operating System | Windows 10 | Ubuntu 18.04 LTS | Windows 7 |
| RAM | 12GB DDR4 | 12GB DDR4 | 2 GB DDR3 |

For each particular setup and JIT state (mixed/interpreted), a matrix **M** was created having number of rows equal to the number of games played and 9 columns. Here the matrix notation $M_{ij}$ indicates the time taken by the algorithm to make the $j^{th}$ move in the $i^{th}$ game; $j$ belongs to the finite set $\{x \mid x \in N, 1 \leq x \leq 9\}$, while $i$ belongs to the finite set $\{x \mid x \in N, 1 \leq x \leq N_g\}$, where $N_g$ is the number of games played.

$$\mathbf{M} = \begin{pmatrix} M_{11} & M_{12} & \cdots & \cdots & M_{19} \\ M_{21} & \ddots & & & M_{29} \\ \vdots & & \ddots & & \vdots \\ \vdots & & & \ddots & \vdots \\ M_{N_g 1} & M_{N_g 2} & \cdots & \cdots & M_{N_g 9} \end{pmatrix} \quad (2)$$

For comparing clock-time analyses across the different setups, this paper formulates the following new notations:

**a. Time per move (TPM)**: It refers to the response time of each move made by the bot. TPM is defined as the average of all moves of a certain move index $j$, for a particular setup, and state of the JIT compiler. TPM gives one an idea of how the algorithm is performing at each move. TPM for the $j^{th}$ move is represented by $TPM_j$, and is given by

$$TPM_j = \frac{\sum_{i=1}^{N_g} M_{ij}}{N_g} \quad (3)$$

**b. Time per game (TPG)**: Time per game is another metric which compares each algorithm on the basis of how long all the nine moves take to execute per game, it refers to the total time taken by each algorithm, on average, to play a game from start to finish. TPG for the $i^{th}$ game is represented by $TPG_i$, and is given by

$$TPG_i = \sum_{j=1}^{9} M_{ij} \quad (4)$$

TPG, the mean of all $TPG_i$ values, is a very good estimate of the speed of an algorithm, and on a certain setup, it is given by

$$TPG = \frac{\sum_{i=1}^{N_g} TPG_i}{N_g} = \frac{\sum_{i=1}^{N_g} \sum_{j=1}^{9} M_{ij}}{N_g} \quad (5)$$

Runtimes are calculated for each move on the three setups with JIT compiler enabled and disabled, using the four algorithms being compared. These 24 matrices are then used to calculate TPMj and TPG, which are used to analyze the performance of the algorithms.

TPG values are tabulated, and the effective speedup of T3DT over the competing algorithms are calculated. Speedup is chosen as the benchmarking criterion here because, keeping all other variables (like setup and JIT state) constant, it gives a true representation of each algorithm's performance. Speedup of T3DT over a certain algorithm X, on a particular setup and JIT state, is given by

$$Speedup\ of\ T3DT\ (X) = \frac{TPG\ (X)}{TPG\ (T3DT)} \quad (6)$$

## 6. Analysis and Results

In this section the claims of improved runtime are supported by irrefutable empirical evidence. Both theoretical as well as practical clock-time analyses have been done for the algorithms as described below.

### 6.1. Theoretical Analysis

This section explores some of the competing approaches to solve this problem and compares and contrasts the merits and demerits of each. The simplest approach for maximizing win-to-draw ratio while still maintaining O(1) time complexity would have been to generate all the possible moves and store them in a tree or directed acyclic graph, and map the current board state to the next optimal state(s). Randomization could be achieved by mapping the current state to a list of optimal next states and selecting one out of them, but this method comes with a high space complexity as it has to store all the sub-trees for the board, maybe even copies of similar board states, in different sub-trees. Hence this approach was discarded.

However, T3DT based on decision trees, has both O(1) space and time complexity, for each move on a fixed 3x3 board, as it stores almost nothing in memory, just the board and about dozen integer values. To make a move, minimax would have to construct the entire game tree at each state, which is quite enormous at the beginning. In contrast, T3DT only has to take a few fixed decisions at every step. In some cases, it might not need to analyse the complete board even once to make a move. This leads to constant space and time complexity in practice.

Some more properties like completeness and optimality of the game tree have been analysed for these algorithms. A search algorithm is said to be complete if an algorithm is guaranteed to reach the goal, provided it exists.

All four algorithms are complete by this definition. Optimality refers to the algorithm reaching in the goal in the least number of moves. Note that completeness is not a guarantee for optimality. Optimality for these algorithms has been discussed in detail in Sec. 3 of this paper.

The asymptotic complexities and some other properties like complete search and optimality for each of the four algorithms being compared MM, ABP, ABA and T3DT are tabulated in Table 6.1. Here for the minimax-based algorithms, complexity is expressed in terms of b and m, where b is the number of legal moves at each point and m is the depth of the tree.

***Table 6.1***: *Theoretical Comparison of the four algorithms*

| Property | MM | ABP | ABA | T3DT |
|---|---|---|---|---|
| Complete Search? | Yes | Yes | Yes | Yes |
| Optimal for optimal opponent? | Yes | Yes | Yes | Yes |
| Optimal for sub-optimal opponent? | No | No | Yes | Yes |
| Time Complexity | $O(b^m)$ | $O(b^{m/2})$ | $O(b^{m/2})$ | $O(1)$ |
| Space Complexity | O(bm) for Depth First Search | | | $O(1)$ |

### 6.2. Practical Clock-time Analysis

This section delves into the actual runtime data collected from running each algorithm against itself. The methodology for the same is discussed in Sec. 5 of this paper. The results are analysed at a granular level, elaborating why and in which moves T3DT outperforms minimax and similar algorithms.

The time taken per move on the three setups are presented in the $TPM_j$ graphs in Figs. 6.1 - 6.3. They show how each algorithm compares against each other for every move. Both linear and semi-log graphs have been used because the linear graphs highlight the significance of the time taken to make the first few moves, while the semi-log scale gives a clearer view of which algorithm performs better in the last few moves, when the magnitude of difference is negligible. The exact time taken for each move from 1 to 9 is also presented in Tables 6.2 – 6.4 for reference and further analysis.

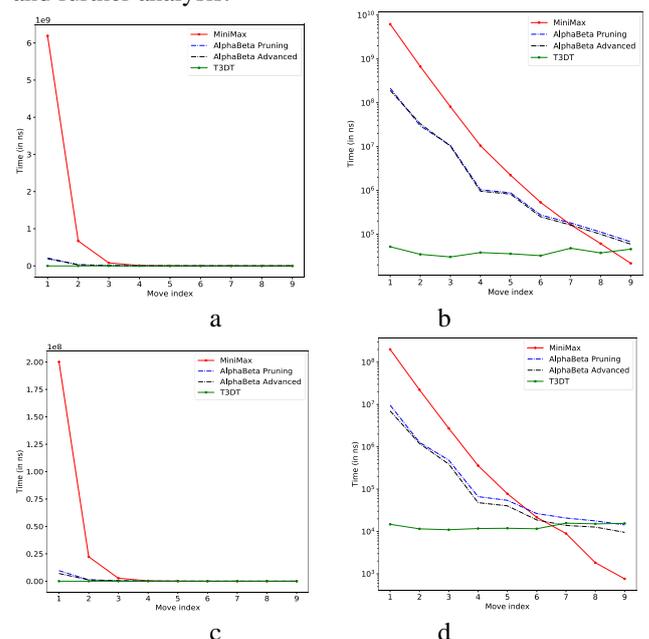

***Fig. 6.1***: $TPM_j\ for\ Setup\ 1$
$(a) linear\ scale,\ no\ JIT\ (b) semi\text{-}log\ scale,\ no\ JIT$
$(c) linear\ scale\ with\ JIT (d) semi\text{-}log\ scale\ with\ JIT$

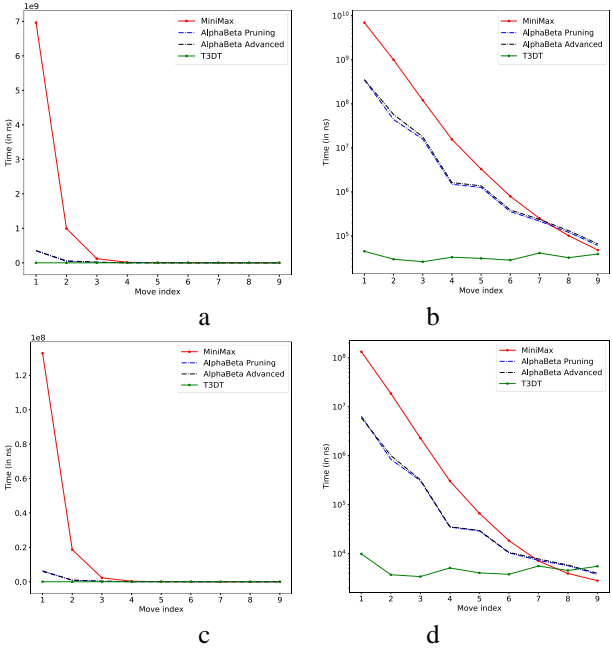
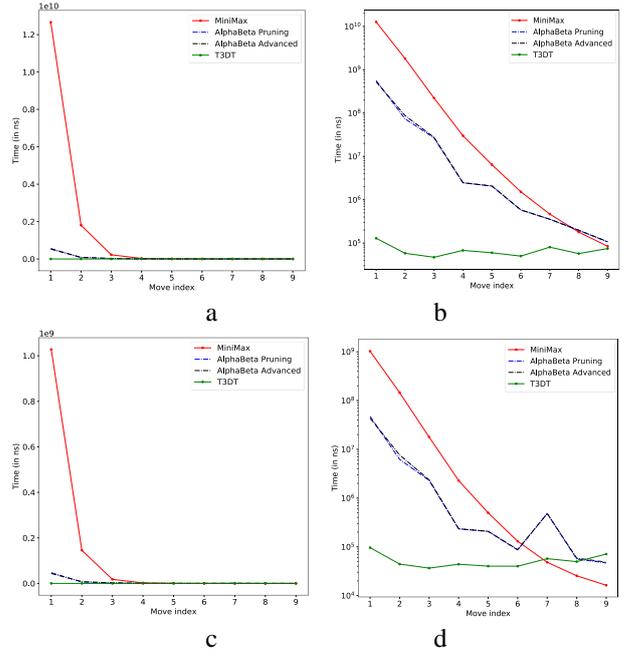

***Fig. 6.2***: $TPM_j$ *for Setup 2*
*(**a**)linear scale, no JIT (**b**)semi-log scale, no JIT*
*(**c**)linear scale with JIT(**d**)semi-log scale with JIT*

***Fig. 6.3***: $TPM_j$ *for Setup 3*
*(**a**)linear scale, no JIT (**b**)semi-log scale, no JIT*
*(**c**)linear scale with JIT(**d**)semi-log scale with JIT*

***Table 6.2***: *TPM values averaged over 10,000 moves for each algorithm on setup 1*

| Algorithm | 1 | 2 | 3 | 4 | 5 | 6 | 7 | 8 | 9 |
|---|---|---|---|---|---|---|---|---|---|
| MM w/o JIT compiler | 6.19E+09 | 6.72E+08 | 8.13E+07 | 1.05E+07 | 2.23E+06 | 5.34E+05 | 1.65E+05 | 6.10E+04 | 2.17E+04 |
| ABP w/o JIT compiler | 2.14E+08 | 2.97E+07 | 1.06E+07 | 1.04E+06 | 8.83E+05 | 2.76E+05 | 1.82E+05 | 1.12E+05 | 6.71E+04 |
| ABA w/o JIT compiler | 1.91E+08 | 3.31E+07 | 1.03E+07 | 9.61E+05 | 8.25E+05 | 2.51E+05 | 1.63E+05 | 1.00E+05 | 5.95E+04 |
| T3DT w/o JIT compiler | 5.19E+04 | 3.49E+04 | 3.04E+04 | 3.83E+04 | 3.59E+04 | 3.25E+04 | 4.80E+04 | 3.76E+04 | 4.58E+04 |
| MM | 2.00E+08 | 2.23E+07 | 2.73E+06 | 3.59E+05 | 7.78E+04 | 2.16E+04 | 8.95E+03 | 1.83E+03 | 7.61E+02 |
| ABP | 9.66E+06 | 1.28E+06 | 4.83E+05 | 6.65E+04 | 5.45E+04 | 2.65E+04 | 2.08E+04 | 1.78E+04 | 1.43E+04 |
| ABA | 7.05E+06 | 1.20E+06 | 3.90E+05 | 4.80E+04 | 4.04E+04 | 1.87E+04 | 1.39E+04 | 1.27E+04 | 9.53E+03 |
| T3DT | 1.48E+04 | 1.15E+04 | 1.10E+04 | 1.18E+04 | 1.19E+04 | 1.16E+04 | 1.58E+04 | 1.52E+04 | 1.56E+04 |

***Table 6.3***: *TPM values averaged over 10,000 moves for each algorithm on setup 2*

| Algorithm | 1 | 2 | 3 | 4 | 5 | 6 | 7 | 8 | 9 |
|---|---|---|---|---|---|---|---|---|---|
| MM w/o JIT compiler | 6.96E+09 | 9.96E+08 | 1.21E+08 | 1.56E+07 | 3.31E+06 | 7.96E+05 | 2.51E+05 | 1.01E+05 | 4.79E+04 |
| ABP w/o JIT compiler | 3.49E+08 | 4.40E+07 | 1.58E+07 | 1.49E+06 | 1.26E+06 | 3.57E+05 | 2.18E+05 | 1.21E+05 | 6.12E+04 |
| ABA w/o JIT compiler | 3.54E+08 | 5.72E+07 | 1.79E+07 | 1.63E+06 | 1.38E+06 | 3.89E+05 | 2.38E+05 | 1.32E+05 | 6.64E+04 |
| T3DT w/o JIT compiler | 4.49E+04 | 2.97E+04 | 2.60E+04 | 3.29E+04 | 3.11E+04 | 2.83E+04 | 4.10E+04 | 3.21E+04 | 3.90E+04 |
| MM | 1.33E+08 | 1.86E+07 | 2.29E+06 | 3.04E+05 | 6.64E+04 | 1.84E+04 | 7.09E+03 | 3.92E+03 | 2.80E+03 |
| ABP | 6.39E+06 | 8.44E+05 | 3.08E+05 | 3.48E+04 | 2.89E+04 | 1.04E+04 | 7.24E+03 | 5.67E+03 | 3.75E+03 |
| ABA | 5.96E+06 | 1.01E+06 | 3.21E+05 | 3.53E+04 | 2.98E+04 | 1.06E+04 | 7.78E+03 | 5.85E+03 | 3.95E+03 |
| T3DT | 9.85E+03 | 3.70E+03 | 3.39E+03 | 5.08E+03 | 4.03E+03 | 3.78E+03 | 5.56E+03 | 4.52E+03 | 5.50E+03 |

***Table 6.4***: *TPM values averaged over 10,000 moves for each algorithm on setup 3*

| Algorithm | 1 | 2 | 3 | 4 | 5 | 6 | 7 | 8 | 9 |
|---|---|---|---|---|---|---|---|---|---|
| MM w/o JIT compiler | 1.27E+10 | 1.81E+09 | 2.22E+08 | 3.00E+07 | 6.43E+06 | 1.53E+06 | 4.67E+05 | 1.79E+05 | 8.45E+04 |
| ABP w/o JIT compiler | 5.62E+08 | 7.35E+07 | 2.68E+07 | 2.47E+06 | 2.08E+06 | 5.81E+05 | 3.54E+05 | 1.97E+05 | 1.07E+05 |
| ABA w/o JIT compiler | 5.24E+08 | 8.73E+07 | 2.76E+07 | 2.46E+06 | 2.07E+06 | 5.83E+05 | 3.55E+05 | 1.98E+05 | 1.08E+05 |
| T3DT w/o JIT compiler | 1.29E+05 | 5.83E+04 | 4.71E+04 | 6.79E+04 | 5.98E+04 | 5.01E+04 | 8.06E+04 | 5.73E+04 | 7.48E+04 |
| MM | 1.03E+09 | 1.46E+08 | 1.79E+07 | 2.29E+06 | 4.98E+05 | 1.30E+05 | 4.82E+04 | 2.55E+04 | 1.63E+04 |
| ABP | 4.73E+07 | 6.22E+06 | 2.30E+06 | 2.34E+05 | 2.07E+05 | 8.61E+04 | 4.83E+05 | 5.62E+04 | 4.67E+04 |
| ABA | 4.36E+07 | 7.65E+06 | 2.37E+06 | 2.34E+05 | 2.08E+05 | 8.79E+04 | 4.81E+05 | 5.86E+04 | 4.84E+04 |
| T3DT | 9.63E+04 | 4.42E+04 | 3.65E+04 | 4.39E+04 | 4.01E+04 | 4.00E+04 | 5.72E+04 | 4.97E+04 | 7.10E+04 |

Quite as expected, minimax takes the highest time at the first step as it has to go through the entire game tree, which comprises of 255,168 unique games. It is followed by minimax with alpha-beta pruning and advanced alpha-beta both of which have similar timings. Initially, T3DT used by the bot has the best response time by a large margin, compared to the other algorithms. Gradually this difference decreases, with decrease in search space and move generation, and in the last few moves, minimax-based approaches might have a faster response than the bot. However, this is insignificant compared to the enormous time taken by these minimax-based approaches in the beginning. For reference, time for the first move constitutes about 80-90% approx. of the total time per game while the time for the last move is approx. 0% of TPG for MM, ABP and ABA. T3DT being a constant time algorithm takes about the same time to make every move.

Furthermore, all the three minimax-based approaches are not randomized, which leads to all the moves made at the same board state by each algorithm being always same. The aim of this paper has been to create a randomized algorithm that also trying to maximize win-to-draw ratio. If a simple non-randomized no-loss decision tree algorithm had been implemented, it is highly probable that the decision tree approach would have been the most time efficient at every move. This is because the randomization in T3DT was observed to take a significant amount of time. Also, in board states where it is easy for the bot to just force a draw and then move in the first empty square till the end thus saving time, it still explores the possibility to win against a sub-optimal player, not only increasing the win-to draw ratio, but also making the game more akin to a human and fun to play with. All these enhancements are novel, but time-consuming, however it only becomes noticeable in the last few moves, when minimax itself takes comparably less time.

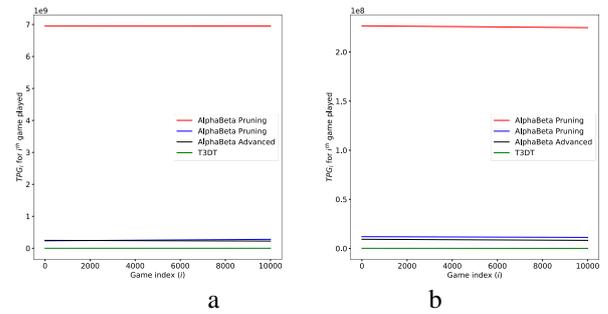

*Fig. 6.4*: *Best fit lines of $TPG_i$ on Setup 1 (a) JIT disabled (b) JIT enabled*

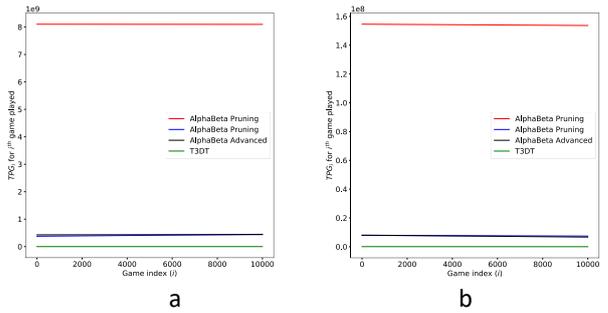

*Fig. 6.5*: *Best fit lines of $TPG_i$ on Setup 2 (a) JIT disabled (b) JIT enabled*

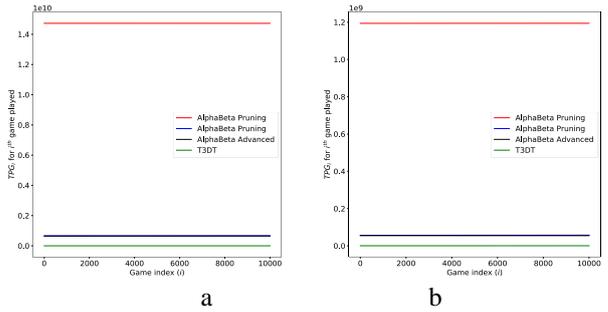

*Fig. 6.6*: *Best fit lines of $TPG_i$ on Setup 3 (a) JIT disabled (b) JIT enabled*

*Table 6.5 TPG for the algorithms on 3 setups*

| Algorithm | Setup 1 | | Setup 2 | | Setup 3 | |
|---|---|---|---|---|---|---|
| | JIT disabled | JIT enabled | JIT disabled | JIT enabled | JIT disabled | JIT enabled |
| MM | 6.96E+09 | 2.26E+08 | 8.10E+09 | 1.54E+08 | 1.47E+10 | 1.19E+09 |
| ABP | 2.57E+08 | 1.16E+07 | 4.13E+08 | 7.64E+06 | 6.68E+08 | 5.69E+07 |
| ABA | 2.36E+08 | 8.78E+06 | 4.33E+08 | 7.38E+06 | 6.45E+08 | 5.47E+07 |
| T3DT | 3.55E+05 | 1.19E+05 | 3.05E+05 | 4.54E+04 | 6.25E+05 | 4.79E+05 |

*Table 6.6*: *Std. Dev of TPG for the algorithms on different setups*

| Algorithm | Setup 1 | | Setup 2 | | Setup 3 | |
|---|---|---|---|---|---|---|
| | JIT disabled | JIT enabled | JIT disabled | JIT enabled | JIT disabled | JIT enabled |
| MM | 2.61E+08 | 3.06E+07 | 9.25E+07 | 1.95E+07 | 4.26E+07 | 1.05E+08 |
| ABP | 3.81E+07 | 4.97E+06 | 4.20E+07 | 2.39E+06 | 4.66E+06 | 4.12E+07 |
| ABA | 2.61E+07 | 3.65E+06 | 1.41E+07 | 2.38E+06 | 6.86E+06 | 4.11E+07 |
| T3DT | 1.55E+05 | 2.39E+05 | 6.58E+04 | 2.34E+05 | 1.07E+06 | 1.91E+06 |

*Table 6.7*: *Speedup of T3DT: No. of times it is faster than other methods*

| Speedup with T3DT | Setup 1 | | Setup 2 | | Setup 3 | | **Average** | |
|---|---|---|---|---|---|---|---|---|
| | JIT disabled | JIT enabled | JIT disabled | JIT enabled | JIT disabled | JIT enabled | **JIT disabled** | **JIT enabled** |
| Minimax | 1.96E+04 | 1.89E+03 | 2.66E+04 | 3.39E+03 | 2.36E+04 | 2.49E+03 | **23232** | **2594** |
| αβ Pruning | 7.22E+02 | 9.75E+01 | 1.35E+03 | 1.68E+02 | 1.07E+03 | 1.19E+02 | **1048** | **128** |
| αβ Advanced | 6.65E+02 | 7.37E+01 | 1.42E+03 | 1.63E+02 | 1.03E+03 | 1.14E+02 | **1039** | **117** |

In Figs. 6.4 - 6.6, each figure representing a setup (please refer specifications in Table 5.1), the best fit lines are constructed to fit the scattered data of $TPG_i$ for each of the 10,000 games played with index $i$, for the 4 algorithms being compared. The order in which the games are played don't matter, hence there's not much to infer from the slope of the lines. However, the lines show the trend of $TPG$ for each algorithm on a particular setup and optimization state. Randomized T3DT is observed to perform better than even all the other non-randomized methods. The exact values of $TPG$ in Table 6.5 and standard deviation in Table 6.6 can be used to infer the nature and spread of the $TPG$ values. T3DT is 3 orders of magnitude faster than ABA when JIT is disabled, otherwise it is around 1-2 orders of magnitude faster. It is around 4 orders of magnitude faster than minimax.

Table 6.7 shows the average speedup (see eq. 6) of T3DT over the other algorithms on different setups, with and without compiler optimization. The results obtained are satisfactory and T3DT is about 23,000 times faster than minimax without optimization and about 2500 times faster with optimization. ABA and ABP are about 1000 and 100 times slower than T3DT without and with JIT enabled.

## 7. Conclusion and Further Work

In this paper, a no-loss algorithm for Tic Tac Toe is presented, its algorithm explained in detail, and runtime compared against existing methods based on various factors. It has been satisfactorily proved that this algorithm is superior to existing minimax-based methods to play Tic Tac Toe. It would be useful to try to generate a generalized rule-based approach like this, creating rules through other methods and then utilizing them for playing games of this class but of higher dimensionality. In its current form, the algorithm can be used to improve decision making in heuristic solutions and minimax-based solutions to play Tic Tac Toe. These rules can also be further analysed to create better fitness function, playing both as the first and second player. Also, it will be interesting to find the exact percentage of victories T3DT achieves out of all possible games of tic tac toe in which victory is achievable, including sub-optimal moves by the opponent.

## 8. Acknowledgements

This work benefitted from discussions with Dr P. Murali, who provided valuable feedback during the final drafting of the paper, his support is gratefully acknowledged.